\pdfoutput=1
\documentclass[twoside,11pt]{article}

\usepackage{jmlr2e}
\usepackage[german,english]{babel}

\providecommand{\tabularnewline}{\\}

\ShortHeadings{Truecluster matching}{Oehlschl\"agel}
\firstpageno{1}

\begin{document}

\title{Truecluster matching}

\author{\name Jens Oehlschl\"agel \email jens.oehlschlaegel@truecluster.com }

\editor{}

\maketitle

\begin{abstract}
Cluster matching by permuting cluster labels is important in many clustering contexts such as cluster validation and cluster ensemble techniques. The classic approach is to minimize the euclidean distance between two cluster solutions which induces inappropriate stability in certain settings. Therefore, we present the $truematch$ algorithm that introduces two improvements best explained in the crisp case. First, instead of maximizing the trace of the cluster crosstable, we propose to maximize a $\chi^2$-transformation of this crosstable. Thus, the trace will not be dominated by the cells with the largest counts but by the cells with the most non-random observations, taking into account the marginals. Second, we suggest a probabilistic component in order to break ties and to make the matching algorithm truly random on random data. The truematch algorithm is designed as a building block of the truecluster framework and scales in polynomial time. First simulation results confirm that the truematch algorithm gives more consistent truecluster results for unequal cluster sizes. Free R software is available. 
\end{abstract}

\begin{keywords}
  Hungarian method, truematch, truecluster, MMCC, CIC, \citet{Hornik:2005}

\end{keywords}

\section{Introduction}

Applying a cluster algorithm to a dataset results in---fuzzy or crisp---assignments of cases to anonymous clusters. In order to interpret these clusters, we often wish to compare these clusters to other classifications, so some heuristic is needed to match one classification to another. With the advent of resampling and ensemble methods in clustering \citep{GordonVichi:2001, DimitriadouWeingesselHornik:2002, StrehlGhosh:2002}, the task of matching cluster solutions has become even more important: we need reliable and scalable matching algorithms that do the task fully automated.

Consider, for example, the use of bootstrapping or cross-validation for cluster validation as suggested by many authors \citep{MoreauJain:1987,JainMoreau:1988,TibshiraniWaltherBotsteinBrown:2001,RothLangeBraunBuhmann:2002,Ben-HurElisseeffGuyon:2002,DudoitFridlyand:2002}: many cluster solutions are created and agreement between them is evaluated. Some agreement indices do not need explicit cluster matching \citep{Rand:1971,HubertArabie:1985}, but others can only be applied \emph{after} cluster solutions have been matched, for example, Cohen's kappa \citeyearpar{Cohen:1960}. 

Recently, authors have suggested transfering the idea of bagging \citep{Breiman:1996} to clustering. Some approaches aggregate cluster centers \citep{Leisch:1999, DolnicarLeisch:2000, Bakker:2001} or aggregate consensus between pairs of observations \citep[\emph{BagClust2} algorithm]{MontiTamayoMesirovGolub:2003, DudoitFridlyand:2003}. Other approaches aggregate cluster assignments and, therefore, require cluster matching, for example, the crisp \emph{BagClust1} algorithm of \citet{DudoitFridlyand:2003}, the combination scheme for fuzzy clustering of \citet{DimitriadouWeingesselHornik:2002} or \emph{truecluster} \citep{Oehlschlaegel:2007a}. 

Truecluster is an algorithmic framework for robust scalable clustering with model selection that combines the idea of bagging with information theoretical model selection along the lines of $AIC$ \citep{Akaike:1973,Akaike:1974} and $BIC$ \citep{Schwarz:1978}. In order to calculate its \emph{cluster information criterion} ($CIC$), truecluster requires a reliable cluster matching algorithm. The truematch algorithm presented here was designed to play that role. The organization of the paper is as follows: in Section \ref{sec:problem}, we show an undesirable feature of the standard approach to cluster matching. In Section \ref{sec:truematch}, we present the truematch algorithm. In Section \ref{sec:truecluster}, we demonstrate the benefits of the truematch algorithm within the truecluster framework. In Section \ref{sec:simulation}, we use simulation to compare truematch against standard trace maximization matching and in Section \ref{sec:discussion}, we discuss our results. 

\section{What's wrong with trace maximization of the matching table\label{sec:problem}}

The standard aproach to cluster matching is searching for that permutation of cluster labels that minimizes the euclidean distance to a reference cluster solution. This criterion has been suggested for fuzzy consensus clustering \citep{GordonVichi:2001, DimitriadouWeingesselHornik:2002}, as well as for crisp consensus clustering \citep{StrehlGhosh:2002} or crisp cluster bagging \citet[BagClust1]{DudoitFridlyand:2003}. In the crisp case, this criterion is simply trace maximization of matching table counts: cross-tabulating class memberships of two solutions and then permuting rows/columns of the matching table until the trace becomes maximal. To our knowledge, cluster publications and software differ in the algorithms used to obtain trace maximization, but do not question the euclidean criterion per se. 

For example, \citet{DimitriadouWeingesselHornik:2002} suggested a recursive heuristic to approximate trace maximization. It is known that trying all permutations has time complexity $\textbf{O}(K!)$, where $K$ denotes the number of clusters. The \emph{Hungarian method} improves on this and achieves polynomial time complexity $\textbf{O}(K^{3})$. \citet{Kuhn:1955} published a pencil and paper version, which was followed by J.R. Munkres' executable  version \citep{Munkres:1957} and extended to non-square matrices by \citet{BourgeoisLassalle:1971}. For a list of further algorithmic approaches to this so-called \emph{linear sum assignment problem} or \emph{weighted bipartite matching}, see \citet{Hornik:2005}. 

However, scalablility is not the only quality aspect of a matching algorithm. An important statistical feature of a matching algorithm is the following: if we match two random partitions, the matching algorithm should not systematically align the two partitions. We now show that the classic trace maximization does not generally possess this feature. 

Assume a cluster algorithm that claims to identify an outlier in a sample of size $N=100$ but which actually declares one case as `outlying' by random. Now assume a procedure that draws two bootstrap samples and clusters them into 99\% `normal' cases and one `outlier'. In 1\% of such procedures, the outlier picked in the second sample will randomly match the outlier picked in the first sample. In such cases, trace maximization matching will lead to a matching table as shown in Table \ref{cap:RandomMatching}. In the other 99\%, there will be no match, which---by trace maximization---gives a matching table like that shown in Table \ref{cap:TraceMaxMatching}. The resulting \emph{expected} matching table is shown in Table \ref{cap:TraceMaxExpected}. 

\begin{table}[ht]
\hfill
   \begin{center}
      \begin{tabular}{c||c|c}
         & a & b \tabularnewline
         \hline 
         \hline 
         a & 99 & 0 \tabularnewline
         \hline 
         b & 0 & 1 \tabularnewline
      \end{tabular}
   \end{center}

   \caption{Random matching (1\%)\label{cap:RandomMatching}}


\hfill
   \begin{center}
      \begin{tabular}{c||c|c}
         & a & b \tabularnewline
         \hline 
         \hline 
         a & 98 & 1 \tabularnewline
         \hline 
         b & 1 & 0 \tabularnewline
      \end{tabular}
   \end{center}

   \caption{Typical trace maximization matching (99\%)\label{cap:TraceMaxMatching}}


\hfill
   \begin{center}
      \begin{tabular}{c||c|c}
         & a & b \tabularnewline
         \hline 
         \hline 
         a & 98.01\% & 0.99\% \tabularnewline
         \hline 
         b & 0.99\% & 0.01\% \tabularnewline
      \end{tabular}
   \end{center}

   \caption{Expected trace maximization matching\label{cap:TraceMaxExpected}}

\end{table}


We can see that under random clustering, we expect 98.02\% on the main diagonal which at first glance looks like a strong (non-random) match. Only applying standard random correction \citep{Cohen:1960} confirms this to be a pure random match (Cohen's kappa = 0). However, in a clustering context we have two objections against relying on such random corrections: as far as evaluation of cluster agreement is concerned, random corrections, such as Cohen's kappa or Hubert and Arabie's corrected rand index do not work properly, because spatial neighbors have an above-random chance of being clustered together in the absence of any cluster structure in the data. Therefore, agreement indices are too optimistic even with random correction. More importantly, in other contexts such as bagging there is no random correction available at all. If cluster sizes are (very) different, bagging cluster results will suffer because in standard trace maximization big randomly matched cells win over small cells representing non-random matches. Therefore, we are looking for a matching algorithm that does not systematically generate a strong diagonal under random conditions.

\section{Truematch algorithm\label{sec:truematch}}

The problems with standard trace maximization described in the previous section result from focusing on raw counts in a situation with unequal marginal (cluster) probabilities. From other contexts, we know that this is not a good idea. Take the $\chi^2$-test for statistical independence of two categorial variables. It is not based on raw counts. Instead, the matching table of raw counts is transformed to another unit taking the marginals into account. Let $N$ denote the total number of observations, $n_{k}$ the number of observations in one row, $n_{l}$ the number of observations in one column and, finally, let $n_{k,l}$ denote the number of observations in one cell of the $K$ x $K$ cluster crosstable. The first step in calculating $\chi^2$ is to calculate for each cell the number of expected counts $\hat{n}_{k,l}$ under the assumption of independence:

\begin{equation} \hat{n}_{k,l} = p_{k}\cdot p_{l}\cdot N=\frac{n_{k}\cdot n_{l}}{N} \label{eq:ExpectedCounts}\end{equation}

Then, we transform the matrix of raw counts in Equation \ref{eq:ExpectedCounts} into a matrix of normalized squared deviations $d_{k,l}$ from the null model:

\begin{equation} d_{k,l} = \frac{(n_{k,l}-\hat{n}_{k,l})^{2}}{\hat{n}_{k,l}} \label{eq:NormalizedSquaredDeviation}\end{equation}

The $\chi^2$-value is defined as the sum of Equation \ref{eq:NormalizedSquaredDeviation} over all cells. If we restore the sign in Equation \ref{eq:NormalizedSquaredDeviation}, we get:

\begin{equation} s_{k,l} = sign(n_{k,l}-\hat{n}_{k,l}) \cdot d_{k,l}  \label{eq:SignedNormalizedSquaredDeviation}\end{equation}

In order to cope with unequal cluster sizes, we suggest basing cluster matching on maximizing the trace of $s_{k,l}$ rather than on maximizing the trace of $n_{k,l}$. And in order to avoid any systematic not based on the data, we add a probabilistic component to the matching algorithm. Consequently we define the \emph{truematch algorithm} as: 

\begin{enumerate}
\item Randomly permute rows and columns of the matching table
\item Transform the matching table counts $n_{k,l}$ to signed normalized squared deviations $s_{k,l}$ using Equation \ref{eq:SignedNormalizedSquaredDeviation}
\item Apply a trace maximization algorithm like the Hungarian method to maximize the trace (in fact the Hungarian method minimizes $-s_{k,l}$)
\item Order the resulting row/column pairs descending by $s_{k,l}$ breaking ties at random
\end{enumerate}

If no trace maximization algorithm like the Hungarian method is available, the matching can easily be done using the \emph{truematch heuristic} similar to the heuristic suggested by \citet{DimitriadouWeingesselHornik:2002}:

\begin{enumerate}
\item Calculate signed normalized squared deviations $s_{k,l}$ for all \emph{remaining} cells of the matching table
\item Order all cells descending by $s_{k,l}$ and by $n_{k,l}$ (breaking ties by random) and denote the first cell as the \emph{target cell}
\item Match the row of the target cell to the column of the target cell 
\item Remove the row and the column of the target cell from the matching table 
\item If both the number of remaining rows and columns is at least two, repeat from step 1
\end{enumerate}

It is obvious that the truematch algorithm has runtime complexity $\textbf{O}(K^{3})$ like the Hungarian method. The truematch heuristic also nicely translates into polynomial runtime. The number of residuals calculated to reduce the matching table from $k$ to $k-1$ is $K^2$, thus the total number of residuals calculated is 
\[
K^2+(K-1)^2+(K-2)^2+...+2^2 = \frac{(K \cdot (K+1))\cdot(2K+1)}{6} - 1
\]
and, therefore, the truematch  heuristic has runtime complexity $\mathbf{O}(K^{3})$ and memory complexity $\mathbf{O}(K^{2})$ if the recursive nature of the algorithm is realized using a while-loop. R package \texttt{truecluster} \citep{R:truecluster} implements the truematch algorithm in \texttt{matchindex(method = "truematch")} and the truematch heuristic in \texttt{matchindex(method = "tracemax")} efficiently through underlying C-code.

Applying the truematch  algorithm and the truematch  heuristic to the above example gives identical results: as in standard trace maximization matching, we find 1\% random matches in matching table \ref{cap:RandomMatching}, but for the 99\% non-random matching cases, truematch generates two versions of matching tables, see Table \ref{cap:TruematchMatching}. Both versions have shifted the majority of counts off-diagonal. Due to the probabilistic component in the 2nd step,  this leads to the expected matching (Table \ref{cap:TruematchExpected}) that has a weak trace. Under truematch, only systematic, non-random matches will result in a strong diagonal.

\begin{table}[ht]
\hfill
   \begin{center}
      \begin{tabular}{c||c|c}
         & a & b \tabularnewline
         \hline 
         \hline 
         a & 1 & 98 \tabularnewline
         \hline 
         b & 0 & 1 \tabularnewline
      \end{tabular}
      ~~~~~~~~~~~~~~~~~~~
      \begin{tabular}{c||c|c}
         & a & b \tabularnewline
         \hline 
         \hline 
         a & 1 & 0 \tabularnewline
         \hline 
         b & 98 & 1 \tabularnewline
      \end{tabular}
   \end{center}
\caption{Typical truematch (49.5\% + 49.5\%)\label{cap:TruematchMatching}}
\end{table}

\begin{table}[ht]
\hfill
   \begin{center}
      \begin{tabular}{c||c|c}
         & a & b \tabularnewline
         \hline 
         \hline 
         a & 1.98\% & 48.51\% \tabularnewline
         \hline 
         b & 48.51\% & 1.00\% \tabularnewline
         \hline
      \end{tabular}
   \end{center}
\caption{Expected truematch \label{cap:TruematchExpected}}
\end{table}

We can quantify the benefit of truematch in this case by comparing expected values of certain agreement indices, cf. Table \ref{cap:Agreement}. The \emph{rand} index \citep{Rand:1971} and its random corrected version \emph{crand} \citep{HubertArabie:1985} are invariant against row/column permutations and, thus, do not differ. There is also no difference for \emph{kappa} \citep{Cohen:1960}. However, the big difference is on the simple non-random-corrected \emph{diagonal} fraction of observations: while the trace maximization misleadingly results in an expected diagonal close to 1, truematch reduces the expectation of this non-random-corrected index close to zero. In the next two sections, we will explore the benefit of truematch in a bagging context, where the main diagonal defines the matching but no random correction is available. 

\begin{table}[ht]
   \hfill
   \begin{center}
      \begin{tabular}{lr|rrrr}
         \hline
         & fraction & diagonal & kappa & rand & crand \tabularnewline 
         \hline
         \hline
         Tracemax RandomMatch & 1.0\% & 1.00 & 1.00&1.000 & 1.00 \tabularnewline
         Tracemax NonRandomMatch & 99.0\% & 0.98 & -0.01 & 0.960 & -0.01 \tabularnewline
         \hline
         Tracemax Expected & 100.0\% & \textbf{0.98} & 0.00 & 0.961 & 0.00 \tabularnewline
         \hline
         \hline
         Truematch Expected & 100.0\% & \textbf{0.03} & 0.01 & 0.961 & 0.00 \tabularnewline
         \hline
         Truematch NonRandomMatch1  & 49.5\% & 0.02 & 0.00 & 0.960 & -0.01 \tabularnewline
         Truematch NonRandomMatch2  & 49.5\% & 0.02 & 0.00 & 0.960 & -0.01 \tabularnewline
         Truematch RandomMatch   & 1.0\% & 1.00 & 1.00 & 1.000 & 1.00 \tabularnewline
         \hline
         \hline
      \end{tabular}
   \end{center}
\caption{Agreement statistics \label{cap:Agreement}}
\end{table}

\section{The role of truematch in truecluster\label{sec:truecluster}}

The truecluster concept \citep{Oehlschlaegel:2007a} suggests a \emph{cluster information criterion} ($CIC$) that evaluates for each cluster model (for each number of clusters) a $N$ x $K$ matrix $\hat{\mathbf{P}}$ that aggregates votes over many resamples. $\hat{\mathbf{P}}$ is created by the \emph{multiple match cluster count} ($MMCC$) algorithm using the truematch algorithm as follows:

\begin{enumerate}
\item Create a $N$ x $K$ matrix $\mathbf{C}$ and initialize each cell $C_{i,k}$
with zero
\item Take a resample (with replacement) of size $N$, use a \emph{base cluster algorithm} to fit the $K$-cluster model $\mathbf{\mathbf{c}^{*}}$ to the resample. Then, use a suitable \emph{prediction method} to determine cluster membership of the out-of-resample cases to get a complete cluster vector \textbf{$\mathbf{c^{'}}$} with $N$ elements $c_{i}^{'}$
\item For each row in $\mathbf{C}$ add one vote (add 1) to the column corresponding to the cluster membership in \textbf{$\mathbf{c^{'}}$}
\item Repeat step 2
\item Estimate cluster memberships $\mathbf{\hat{c}}$ by row-wise majority count in $\mathbf{C}$ (breaking ties at random), use the truematch \emph{algorithm} or \emph{heuristic} to align \textbf{$\mathbf{c^{'}}$} with $\hat{\mathbf{c}}$, and rename the clusters in \textbf{$\mathbf{c^{'}}$} like the corresponding clusters in $\hat{\mathbf{c}}$
\item For each row in $\mathbf{C}$ add one vote (add 1) to the column corresponding to the cluster membership in \textbf{$\mathbf{c^{'}}$}
\item Repeat from step 4 until some reasonable \emph{convergence criterion} is reached
\item Divide each cell in $\mathbf{C}$ by its rowsum to get a matrix of estimated cluster membership probabilities $\hat{\mathbf{P}}$
\end{enumerate}

Table \ref{cap:cluematch} summarizes simulations with truecluster versus consensus clustering: 100 cases, 10,000 replications, for details see \texttt{MMCCconcensus.r} in R package \texttt{truecluster} \citep{R:truecluster}, the table is sorted and grouped by the magnitude of CIC values). For random data without cluster structure, we would expect very `fuzzy' $\hat{\mathbf{P}}$ without clear preferences for any cluster. Furthermore, we would expect CIC to increase for models with more true clusters and to decrease if models try to distinguish more clusters than justified by the data. 

Table \ref{cap:cluematch} shows that the MMCC algorithm using truematch delivers on this expectation: CIC increases for justified clusters and declines for unjustified ones, even if unjustified clusters in the model are small. This works because once cluster decisions are unjustified, the trumatch algorithm starts distributing its votes randomly across undistinguishable columns of $\mathbf{C}$ and, thus, `fuzzifies' $\hat{\mathbf{P}}$. Compare that to consensus clustering \citep{DimitriadouWeingesselHornik:2002} based on trace maximization obtained with R package clue \citep{R:clue, Hornik:2005}. Models with unjustified small clusters get CIC values as high as models without the unjustified cluster. This is a consequence of the trace maximization matching, adding inappropriate stability to the voting. Take, for example, the "random 99:1" model, which is as unjustified as the "random 50:50" model but receives a much higher $CIC$ value. The stability induced by the trace maximization matching results in quite a crisp $\hat{\mathbf{P}}_{2}$: for each row, we find high probability for one cluster and low probability for the other. If we assign cases to clusters based on the maximum probability per row in $\hat{\mathbf{P}}$, all cases are assigned to the same cluster. Such a degenerated $\hat{\mathbf{P}}$ is not wrong but unfortunate. If we manually analyze $\hat{\mathbf{P}}_{2}$, we might detect that $\hat{\mathbf{P}}_{2}$ actually represents a one-cluster (K=1) model. But if we are after automatic selection of models (number of clusters), it is misleading that $\hat{\mathbf{P}}_{2}$ does not represent $K=2$ but $K=1$. Analyzing a consensus cluster solution $\hat{\mathbf{P}}_{K}$ for degeneracies does not really help: the estimated probabilitites can be biased even before the matrix formally degenerates.

\begin{table}[ht]
   \hfill
      \begin{tabular}{lrr|rrrr}
         MMCC & true K & model K & H & RMC & I & CIC \tabularnewline 
         \hline
         \hline
         random 50:49:1       & 1 & 3  & 1.578 & 0.020 & 0.044 & -1.534 \tabularnewline
         \hline
         random 99:1          & 1 & 2  & 1.000 & 0.010 & 0.014 & -0.985 \tabularnewline
         random 50:50         & 1 & 2  & 0.995 & 0.010 & 0.059 & -0.936 \tabularnewline
         \hline
         single 100            & 1 & 1  & 0.000 & 0.000 & 0.000 &  0.000 \tabularnewline
         justified 50 random 49:1 & 2 & 3  & 0.499 & 0.018 & 0.695 &  0.196 \tabularnewline
         \hline
         justified 50:50          & 2 & 2  & 0.000 & 0.010 & 0.990 &  0.990 \tabularnewline
         \hline
         \hline
          &  & & & \tabularnewline 
         consensus & true K & model K & H & RMC & I & CIC \tabularnewline 
         \hline
         \hline
         random 50:49:1       & 1 & 3  & 1.066 & 0.011 & 0.049 & -1.016 \tabularnewline
         random 50:50         & 1 & 2  & 0.995 & 0.010 & 0.048 & -0.947 \tabularnewline
         \hline
         random 99:1          & 1 & 2  & 0.081 & 0.001 & 0.001 & -0.080 \tabularnewline
         single 100            & 1 & 1  & 0.000 & 0.000 & 0.000 &  0.000 \tabularnewline
         \hline
         justified 50 random 49:1 & 2 & 3  & 0.071 & 0.011 & 0.965 &  0.895 \tabularnewline
         justified 50:50          & 2 & 2  & 0.000 & 0.010 & 0.990 &  0.990 \tabularnewline
         \hline
          &  & & & \tabularnewline 
      \end{tabular}

      \begin{tabular}{r|l}
      true K & true number of clusters \tabularnewline
      model K & model number of clusters \tabularnewline
      H & model uncertainty \tabularnewline
      RMC & relative model complexity \tabularnewline
      I& model information \tabularnewline
      CIC & cluster information criterion (I-H) \tabularnewline
      \hline
      single 100 & theoretical values for single group (no cluster) \tabularnewline
      random 50:50 & random clustering with 2 equal sized clusters \tabularnewline
      random 99:1 & random clustering 2 unequal sized clusters \tabularnewline
      random 50:49:1 & random clustering with 3 unequal sized clusters \tabularnewline
      justified 50:50 & justified clustering with 2 equal sized cluster \tabularnewline
      justified 50 random 49:1 & 2 justified clusters, one randomly split unequal sized \tabularnewline
      \end{tabular}

\caption{consensus cluster vs. truecluster \label{cap:cluematch}}

\end{table}

\clearpage

\section{Simulation results\label{sec:simulation}}

In order to systematically investigate the consequences of the different features of truematch versus simple trace maximization matching, we have carried out extensive simulations within the truecluster framework: we assume two clusters and vary their relative size $p$ and the reliability $\kappa$ of a fictitious clustering algorithm and compare the $truecluster$ results gained via trace maximization versus truematch. We did two versions of the simulations: in the \emph{non-fixed} version, $p$ just determines sampling probabilitites; in the \emph{fixed} version, the fictitious clustering algorithm enforces the exact relative size $p$ of the two clusters. Details of the simulation are given in Appendix A. 

Figure \ref{cap:NonfixedSimul} shows \emph{information}, \emph{uncertainty}, and its difference $CIC$ for the non-fixed simulations. White areas denote simulation trials where the truecluster algorithm degenerated from a 2-cluster solution to a 1-cluster solution. The most notable difference is the big share of non-converged truecluster solutions using trace maximization, compared to the truematch algorithm. The estimated information, given reliability and skewness, is very similar and reasonable: information is highest for $p=0.5$ and $\kappa=1.0$ and is lower for both reducing $\kappa$ and/or skewing $p$. 

\begin{figure}[ht]
\includegraphics[%
  width=1.0\columnwidth,
  keepaspectratio,
  angle=-90]{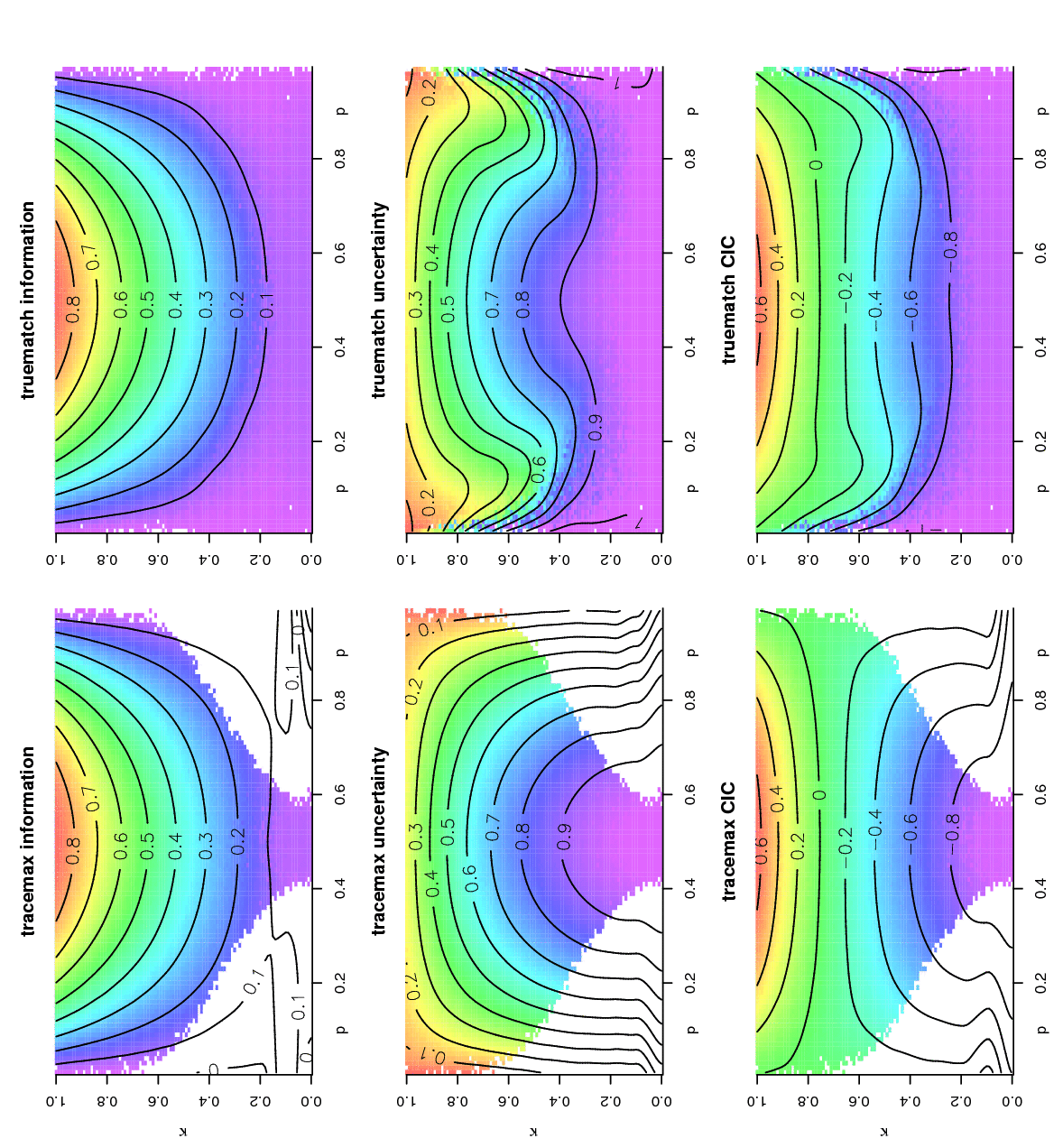}

\caption{Results of fixed simulations \label{cap:FixedSimul}}
\end{figure}

By contrast, compared for uncertainty and for the $CIC$, trace maximization and truematch differ dramatically. Using trace maximization, the uncertainty estimate does not only depend on $\kappa$ but is also artificially lower for higher skewness. As a consequence, cluster models with unequal cluster sizes get better $CIC$ values than cluster models with equal cluster sizes. Using the truematch algorithm almost avoids this undesirable pattern: the estimated uncertainty almost only depends on $\kappa$, not on $p$. The estimated $CIC$ shows a very reasonable pattern: at high $\kappa$ the $CIC$ is highest for equal sized clusters---conforming with the entropy principle--- at low $\kappa$, the $CIC$ is low, however skewed $p$ is. Only at very extreme $p$ is the $CIC$ biased downwards: too small clusters cannot be detected with too small a sample size. Extreme models are non-identifiable and the uncertainty estimate has high variance. Keep in mind that `extreme' $p$ corresponds to very few cases at a sample size of $N=100$. The fixed simulations gave similar results (Figure \ref{cap:FixedSimul}).

\begin{figure}[ht]
\includegraphics[%
  width=1.0\columnwidth,
  keepaspectratio,
  angle=-90]{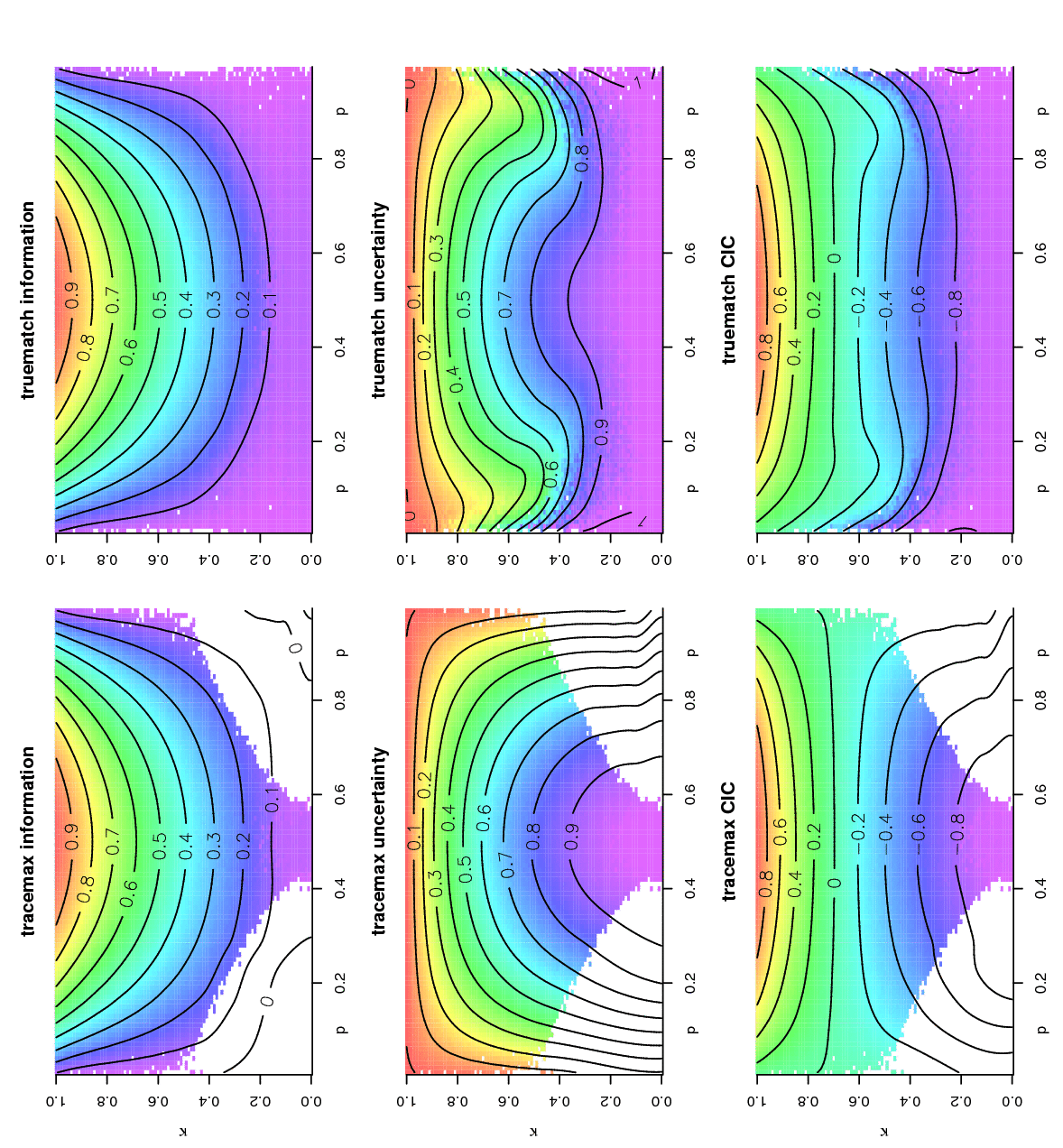}

\caption{Results of non-fixed simulations \label{cap:NonfixedSimul}}
\end{figure}

In summary, trace maximization fails to estimate uncertainty independent of skewness and tends to overestimate $CIC$ for unequal cluster sizes or fails to converge. This restricts its usefulness for cluster evaluation and bagging. By contrast, the truematch algorithm works at almost any combination of reliability and skewness (with the exception of non-identifiable models, given the sample size).

\clearpage

\section{Discussion\label{sec:discussion}}

We have shown that trace maximization matching fails to behave sufficiently neutrally when matching clusterings. The problem arises generally but is especially important in contexts where random correction is not applicable. As an alternative, we have presented the truematch algorithm and heuristic, both probabilistically generate neutral expected matching tables and scale in polynomial time. Our simulations have confirmed that truematch avoids unjustified (expected) matchings induced by unequal cluster sizes. For the simulations done here, the truematch  algorithm and the truematch heuristic behave identically. Since the truematch heuristic does not guarantee maximizing the $\chi^2$-criterion, we expect the truematch algorithm to be superior. However, there is a subtle difference: while the matching of the truematch algorithm depends solely on $s_{k,l}$, the truematch heuristic uses $s_{k,l}$ \emph{and} $n_{k,l}$ to select the row/column matches. Therefore, a final decision about an optimal matching algorithm needs more investigation. 

Truematch is central to the $MMCC$ algorithm, which creates the basis for the CIC-evaluation in the truecluster framework and, thus, contributes to solving the decade-old problem of choosing the optimal number of clusters. Beyond that, cluster bagging, in general, could benefit from using truematch: the resulting  $N$ x $K$ matrix is rather fuzzified than degenerated for unjustified cluster splits. This allows for better automated processing of such results. It is an open question whether the truematch algorithm also has advantages for consensus clustering, or whether different usages of cluster ensembles require different matching algorithms.

\acks{
We would like to thank Dr. Stefan Pilz for reviewing this paper and giving valuable hints for improvement.
}

\newpage

\appendix
\section*{Appendix A.}
\label{app:simulation}

In this appendix, we give details concerning the simulations in section~\ref{sec:simulation}: 
assume a vector $x$ of length 100 with `true' sample group memberships where $p$ denotes the fraction of 1 and $(1-p)$ fraction of 0. 
Let $\mathbf{p}_1$ denote the matrix of joint probabilities for a case's true and clustered classification when the cluster algorithm perfectly separates 0 from 1 (at $\kappa=1$). 

\[ \mathbf{p}_1 = 
\begin{array}{|cc|}
(1-p) & 0\\
0 & p
\end{array}
\]

Let $\mathbf{p}_0$ denote the matrix of joint probabilities for a case's true and clustered classification when the cluster algorithm makes a random guess when separating 0 from 1 (at $\kappa=0$). 

\[ \mathbf{p}_0 = 
\begin{array}{|cc|}
(1-p)^{2} & (1-p) \cdot p\\
(1-p) \cdot p & p^{2}
\end{array}
\]

Then $p_\kappa$ denotes the matrix of joint probabilities for a case's true and clustered classification when the cluster algorithm has reliability $\kappa$. 

\[ p_\kappa = \kappa \cdot \mathbf{p}_1 + (1-\kappa) \cdot \mathbf{p}_0 \]

The two conditional probabilitîes $\mathbf{p}_{id}$ that the clustering algorithm identifies the true class, given the true class, are

\[ \mathbf{p}_{id} = \kappa + (1-\kappa) \cdot 
\begin{array}{|c|}
(1-p) \\
p
\end{array}
\]

\noindent
For each value of $p \in \{1/100, 2/100 .. 99/100\}$ and each value of $\kappa \in \{0.00, 0.01, 0.02, .., 1.00\}$, we simulate aggregation of 1000 bootstrap samples from $x$, for each bootstrap sample our fictitious cluster algorithm assigns cases with probability $\mathbf{p}_{id}$ to the true class and with probability $1-\mathbf{p}_{id}$ to the other class. The resulting cluster memberships $\mathbf{\mathbf{c}^{*}}$ are matched versus the (current) estimated cluster memberships $\mathbf{\hat{c}}$ of the cases in the bootstrap sample. If $\mathbf{\mathbf{c}^{*}}$ or $\mathbf{\hat{c}}$ does not contain two classes, the bootstrap sample is dropped and replaced by another one. Differently from the $MMCC$ algorithm in Section~\ref{sec:truecluster}, we do not predict cluster memberships of the out-of-bag cases. We use $\mathbf{\mathbf{c}^{*}}$ directly instead of \textbf{$\mathbf{c^{'}}$}, consequently the rows of $\mathbf{C}$ are not guaranteed to have aggregated an equal number of votes. For all combinations of $p$ and $\kappa$---the resulting 99x101 truecluster models $\hat{\mathbf{P}}$---we calculate \emph{information}, \emph{uncertainty}, and $CIC$ \citep{Oehlschlaegel:2007a}. These values are visualized using colorcoding and  contourlines are added based on a loess smooth. To create the $fixed$ version, the complete procedure is repeated, additionally enforcing a fixed fraction $p$ by moving randomly selected observations in $\mathbf{\mathbf{c}^{*}}$ from the too big group to the too small one---analogous to a cluster algorithm that forces certain cluster sizes. The R-code doing the simulation is available in \texttt{truematch.r} in package \texttt{truecluster} \citep{R:truecluster}.

\newpage

\bibliography{tc}

\end{document}